\documentclass{bmvc2k}

\usepackage{pifont}
\newcommand{\cmark}{\ding{51}}%
\newcommand{\xmark}{\ding{55}}%
\usepackage{multirow}
\usepackage{booktabs}
\usepackage{amsmath}
\usepackage{algorithm}
\usepackage{algorithmic}
\usepackage{caption}
\usepackage{soul}
\usepackage{framed}
\usepackage{cancel}

\usepackage{blindtext}

\title{Towards Clip-Free Quantized Super-Resolution Networks: How to Tame Representative Images}

\addauthor{Alperen Kalay}{alperenkalay@aselsan.com.tr}{1}
\addauthor{Bahri Batuhan Bilecen}{batuhanb@aselsan.com.tr}{1}
\addauthor{Mustafa Ayazoglu}{mayazoglu@aselsan.com.tr}{1}

\addinstitution{
 ASELSAN Inc.\\
 UGES Division\\
 Ankara, Türkiye
}

\runninghead{Kalay, Bilecen, Ayazoglu}{Towards Clip-Free Quantized SR Networks}

\def\etal{\emph{et al}\bmvaOneDot}

\begin{document}
\maketitle
\vspace{-0.5cm}
\begin{abstract}
Super-resolution (SR) networks have been investigated for a while, with their mobile and lightweight versions gaining noticeable popularity recently. Quantization, the procedure of decreasing the precision of network parameters (mostly FP32 to INT8), is also utilized in SR networks for establishing mobile compatibility. This study focuses on a very important but mostly overlooked post-training quantization (PTQ) step: representative dataset (RD), which adjusts the quantization range for PTQ. We propose a novel pipeline \textbf{(clip-free quantization pipeline, CFQP)} backed up with extensive experimental justifications to cleverly augment RD images by only using outputs of the FP32 model. Using the proposed pipeline for RD, we can successfully eliminate unwanted clipped activation layers, which nearly all mobile SR methods utilize to make the model more robust to PTQ in return for a large overhead in runtime. Removing clipped activations with our method significantly benefits overall increased stability, decreased inference runtime up to 54\% on some SR models, better visual quality results compared to INT8 clipped models - and outperforms even some FP32 non-quantized models, both in runtime and visual quality, without the need for retraining with clipped activation.
\vspace{-0.55cm}
\end{abstract}

\begin{figure}[ht!]
    \centering
    \addtolength{\tabcolsep}{-10pt}
    \begin{tabular}{cc}
    \includegraphics[scale=0.25, trim={0 0cm 0cm 1.8cm}, clip]{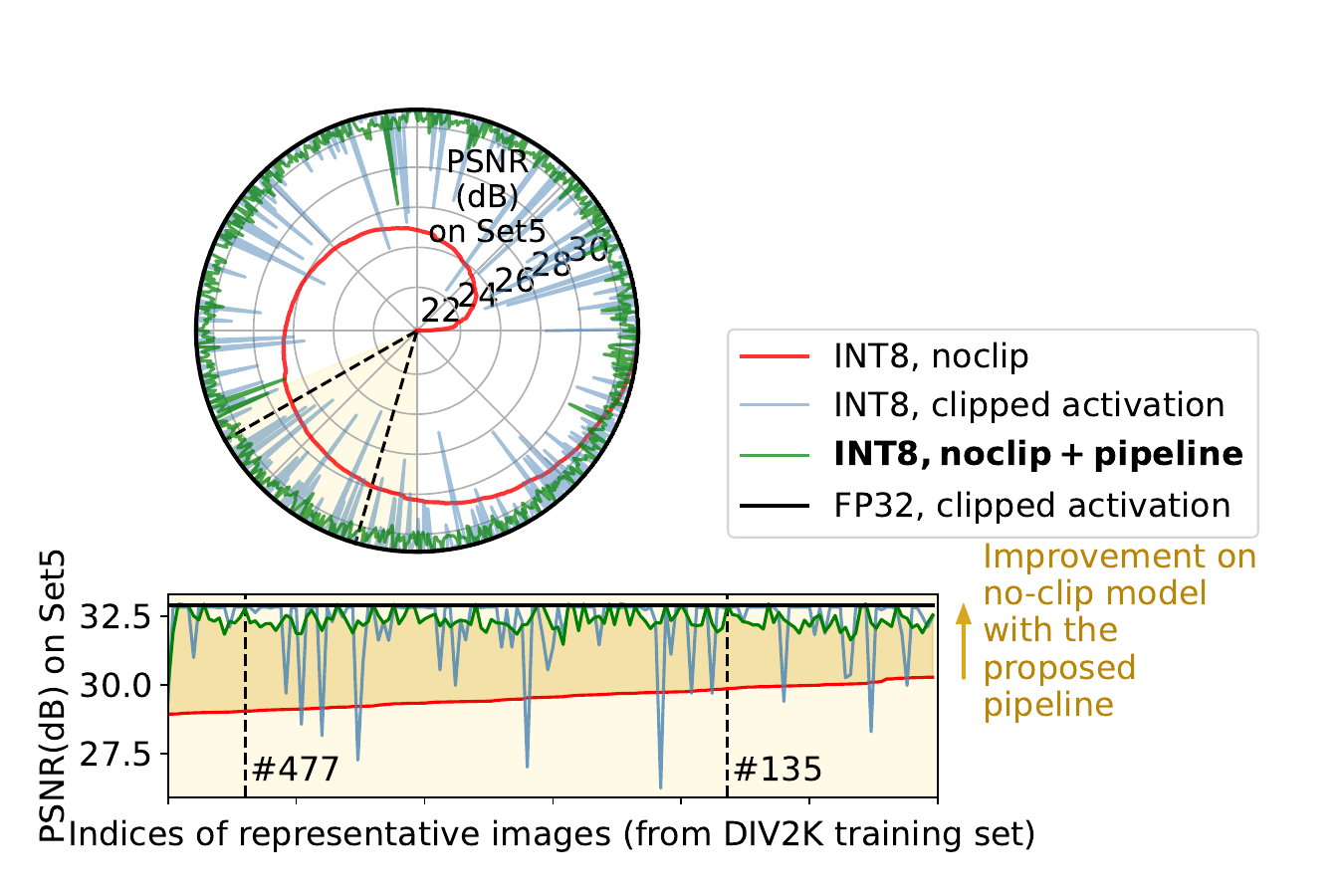} &
    \includegraphics[scale=0.35, trim={1cm 0cm 0cm 0cm}, clip]{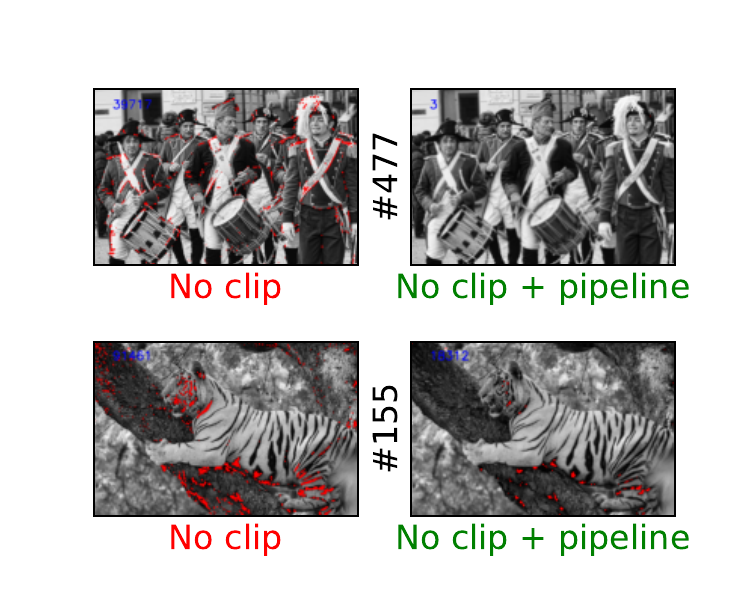} 
    \end{tabular}
    \caption{PSNR on Set5~\cite{set5} for XCAT~\cite{xcat} (radial \& y-axis), where the model is individually quantized with 800 different representative images on DIV2K~\cite{div2k} (angular \& x-axis). We aim to get closer to the circumference. For clarity, we sort INT8 no-clip PSNR values (red) and use the same sorted indices for the rest. The no-clip INT8 XCAT model with CFQP outperforms the same model when RD is not augmented.The no-clip INT8 model with CFQP yields fewer outliers in the output data (right), increasing the visual quality.}
    \label{fig:teaser_circle}
\end{figure}

\begin{figure}[ht!]
    \centering
    \includegraphics[scale=0.80]{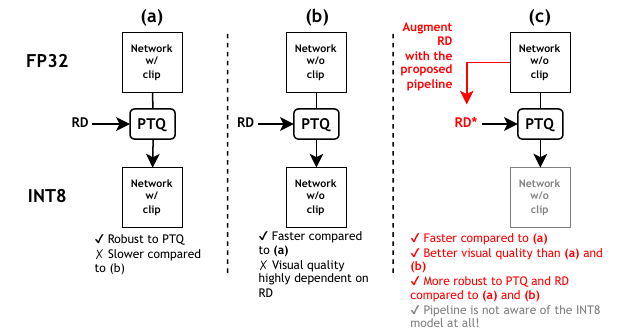}
    \vspace{0.1cm}
    \caption{Most SR networks to be gone under PTQ includes a clipped activation at the end to limit the output range (a). If the clip is removed, the visual quality gets very unstable (i.e., not robust to PTQ) (b), most of the time making the network useless. We propose an RD augmentation pipeline (CFQP, c), which only utilizes the outputs of the FP32 model to be quantized and show significant improvements over (a) \& (b).}
    \label{fig:teaser_flow}
\end{figure}

\section{Introduction}
\label{sec:intro}
Super-resolution (SR) is a widely studied ill-posed inverse problem that involves generating high-resolution (HR) images from low-resolution (LR) ones.  Despite remarkable progress achieved through deep learning on SR networks, the emphasis has been predominantly on enhancing the quality of SR results rather than addressing the efficiency and deployment challenges associated with resource-constrained mobile platforms. One promising technique to address this challenge is post-training quantization (PTQ), which compresses the model size by decreasing the precision of network parameters (commonly from FP32 to INT8), reducing the inference time. 

However, despite its effectiveness, PTQ can significantly drop network performance if not applied correctly. It becomes more critical, especially on low-level computer vision tasks such as SR. Recent advancements in efficient \& mobile SR methods~\cite{xlsr, mai2021, aim2022, esr_ntire_2022,zamfir2023rtsr, conde2023ntire_rtsr} reveal that using a clipped activation layer (mostly clipped ReLU) at the end of the network model aids in keeping the visual quality consistent in FP32 training and PTQ processes, but with a cost of an increase in the inference which is more noticeable on limited hardware. The clipping operation helps PTQ to be more stable because the FP32 model has already been forced to map its output values to the correct range. Hence, quantizing those values would not yield out-of-range results, keeping the output visual quality consistent between INT8 and FP32 models as much as possible.

That said, one very important yet neglected part of these previously followed procedures is that PTQ already determines a range for the quantization internally (hence the clipping range) using the given \textit{representative dataset (RD)}. This means that most of the previous studies have not selected their representative datasets with any optimization, hence having to rely on the additional clipped activation at the network's end to force the PTQ to set its quantization range to whatever the clipped activation layer's range is. Most other works use all the training set, take random subsets, or individual images to construct RD. Still, they do not have an optimized approach and do not aim to eliminate the clipped activations for SR models.

Hence, this paper aims to optimize the overlooked part of choosing RD by performing extensive experiments and observations. Based on our observations, we propose \textbf{clip-free quantization pipeline (CFQP)} for RD to make the INT8 network more robust to PTQ, which eliminates the need for clipped activations to keep the visual quality consistent (Figure~\ref{fig:teaser_circle} and~\ref{fig:teaser_flow}.c). Our pipeline consists of determining \textit{good} and \textit{bad} images for PTQ based on the number of outlier pixel numbers and ranges, and improves the \textit{bad} ones only by utilizing the outputs of the corresponding FP32 model. Our contributions are as follows:

\begin{itemize}
\setlength\itemsep{0.05em}
\item We prove that the RD's for SR yields better performance when they only have a single representative image (RI), eliminating the need for constructing an unnecessarily large RD. This also serves as one-shot quantization.

\item We show that even if the model is trained with RGB images, choosing RD as grayscale yields much higher PSNR scores after PTQ compared to RGB.

\item We determine whether an image is a \textit{good} or a \textit{bad} RI candidate by only looking at the output response of corresponding FP32 models.

\item We propose CFQP to augment and improve \textit{bad} RI's, which improve the overall INT8 model and eliminate the need for clipped activations. The no-clip INT8 models with our modified RI's outperform the same no-clip and clipped INT8 models without augmented RI's, and even some FP32 models in terms of PSNR. With a high ratio (99.1\%), our proposed pipeline makes every \textit{bad} image a \textit{good} candidate for RI.

\item Given the need for clipped activations is alleviated, we achieve significant SR model inference speedups (up to 54\%) compared to their clipped counterparts on various mobile devices.
\end{itemize}

\section{Related Work}
\label{sec:related_work}

Quantization is frequently used and applied on image/video SR models, motivated by the ongoing challenges and related workshops~\cite{aim2022, esr_ntire_2022}. In addition, there are also studies to improve quantization accuracy, like using new quantization-aware loss functions~\cite{LPQ_tinyML} or rearranging activation functions like clipping~\cite{PACT}. On the other hand, there are some notable works on representative dataset (RD) effects. PTQ and RD are considered together in cross-domain calibration studies~\cite{cross-domain}. However, it does not aim to classify good representative images in the training dataset and improve others. Besides the cross-domain, dataset distillation is also a recent field to create a condensed dataset, but generally aimed at training networks and includes no alteration in image content~\cite{dataset-distillation}.
 
Apart from these, Zhang~\etal. proposed a method to select representative images for PTQ~\cite{selectQ}. They use clustering: which may vary for different datasets, does not benefit from the FP32 model output information, and does not try to improve the bad-performing representative images, unlike ours. Yuan~\etal also did some benchmark tests considering PTQ~\cite{benchmark_PTQ}. In a small section, they mention a calibration dataset selection. Still, it does not include a detailed investigation, and we have shown some conflicting results with justifications about the selection of representative dataset size. 

\section{Preliminary Experiments}

\label{sec:preliminary_observations}
In this section, we explain our setup and ask \& answer the relevant questions related to the selection of RD. We have conducted multiple experiments to find satisfying answers to the proposed questions, ultimately leading us to our CFQP proposal.

\subsection{Model Selection and Training}
 We utilize 5 mobile-friendly, diverse in structure \& activations SR models with upscaling rate x3: XCAT~\cite{xcat}, ABPN~\cite{abpn}, ESPCN~\cite{espcn}, FSRCNN~\cite{fsrcnn}, and RFDNNet~\cite{rfdnnet}. The selection of networks and the reason for x3 is to be consistent with the recent and comparable mobile SR challanges~\cite{conde2023ntire_rtsr,aim2022,esr_ntire_2022}. If the clip is enabled, we trained with \textit{clipped ReLU} at the end of the networks. We opted for \textit{TensorFlow (TF)} for training FP32 models and applied INT8 PTQ using \textit{TensorFlow Lite}. We also considered ONNX and PyTorch; however, we stuck with our decision due to TF being the most mature one in model quantization. Models are trained for 400 epochs, each consuming randomly picked 16x16 low-resolution patches from DIV2K~\cite{div2k} training set (800 images) with a batch size of 8. We stick with DIV2K for quantization experiments because we noticed that mobile networks do not benefit from the abundance of training data due to their small number of parameters, although Flickr has nearly thrice of images compared to DIV2K (Table~\ref{tab:tab_flickr}). We use all images of the DIV2K validation set for validation. Adam optimizer is utilized with $\beta_{1,2}$ = 0.99 and 0.999, respectively. The learning rate schedule is as follows: we start the training with 4e-4, linearly increase to 25e-4 in the first ten epochs, and decrease to 4e-5 in the remaining epochs. We employ Set5~\cite{set5} for the test set and measure all PSNR values on the Y channels, compliant with the common convention. For quantization, static-PTQ is used. From now on, PTQ denotes static-PTQ unless specified otherwise.

\subsection{Observations}

\begin{table}[ht!]
\setlength{\tabcolsep}{1.5pt}
\centering
\fontsize{6pt}{6pt}\selectfont
\begin{tabular}{@{}rcl|ll|cl|ll|cl|ll|cl|ll|cl|ll@{}}
\toprule
{Models} & \multicolumn{4}{c}{\textbf{XCAT}}                                                                                    & \multicolumn{4}{c}{\textbf{ABPN}}                                                                                    & \multicolumn{4}{c}{\textbf{ESPCN}}                                                                                   & \multicolumn{4}{c}{\textbf{FSRCNN}}                                                                                  & \multicolumn{4}{c}{\textbf{RFDNNet}}                                                                                 \\ \cmidrule(lr){2-5} \cmidrule(lr){6-9} \cmidrule(lr){10-13} \cmidrule(lr){14-17} \cmidrule(lr){18-21}
Precision                  & \multicolumn{2}{c}{FP32}                          & \multicolumn{2}{c}{INT8}                                & \multicolumn{2}{c}{FP32}                          & \multicolumn{2}{c}{INT8}                                & \multicolumn{2}{c}{FP32}                          & \multicolumn{2}{c}{INT8}                                & \multicolumn{2}{c}{FP32}                          & \multicolumn{2}{c}{INT8}                                & \multicolumn{2}{c}{FP32}                          & \multicolumn{2}{c}{INT8}                                \\ \cmidrule(lr){2-3} \cmidrule(lr){4-5} \cmidrule(lr){6-7} \cmidrule(lr){8-9} \cmidrule(lr){10-11} \cmidrule(lr){12-13} \cmidrule(lr){14-15} \cmidrule(lr){16-17} \cmidrule(lr){18-19} \cmidrule(lr){20-21} 
Clip                       & \cmark               & \multicolumn{1}{c}{\xmark} & \multicolumn{1}{c}{\cmark} & \multicolumn{1}{c}{\xmark} & \cmark               & \multicolumn{1}{c}{\xmark} & \multicolumn{1}{c}{\cmark} & \multicolumn{1}{c}{\xmark} & \cmark               & \multicolumn{1}{c}{\xmark} & \multicolumn{1}{c}{\cmark} & \multicolumn{1}{c}{\xmark} & \cmark               & \multicolumn{1}{c}{\xmark} & \multicolumn{1}{c}{\cmark} & \multicolumn{1}{c}{\xmark} & \cmark               & \multicolumn{1}{c}{\xmark} & \multicolumn{1}{c}{\cmark} & \multicolumn{1}{c}{\xmark} \\ \midrule
$PSNR_{R=RGB}$                       & \multicolumn{1}{l}{\multirow{2}{*}{32.87}} & \multicolumn{1}{l|}{\multirow{2}{*}{32.98}}                           &   32.63                         & 26.57                            & \multicolumn{1}{l}{\multirow{2}{*}{33.13}} & \multicolumn{1}{l|}{\multirow{2}{*}{33.18}}                           &  32.85                          & 26.13                            & \multicolumn{1}{l}{\multirow{2}{*}{32.23}} & \multicolumn{1}{l|}{\multirow{2}{*}{32.23}}                           &  31.18                        & 24.92                            & \multicolumn{1}{l}{\multirow{2}{*}{32.83}} &  \multicolumn{1}{l|}{\multirow{2}{*}{32.83}}                          &  31.35                          & 25.69                            & \multicolumn{1}{l}{\multirow{2}{*}{33.02}} &  \multicolumn{1}{l|}{\multirow{2}{*}{32.97}}                           &   32.23                         & 25.62                            \\
$PSNR_{R=Gray}$                       & &                            &      32.31                       & 28.48                            &  &                            & 32.52                            & 28.44                            & &                           & 30.85                            & 26.26                            &  &                            & 31.12                            & 27.24                            & &                            & 31.76                            & 27.05                            \\ \midrule 
$\Delta_{R=RGB}$ & \multicolumn{2}{c|}{\multirow{2}{*}{-0.11}} &   \multicolumn{2}{c|}{6.06} & \multicolumn{2}{c|}{\multirow{2}{*}{-0.05}} & \multicolumn{2}{c|}{6.72} & \multicolumn{2}{c|}{\multirow{2}{*}{$\sim$0}} & \multicolumn{2}{c|}{6.26} & \multicolumn{2}{c|}{\multirow{2}{*}{$\sim$0}} & \multicolumn{2}{c|}{5.66} &  \multicolumn{2}{c|}{\multirow{2}{*}{-0.05}} & \multicolumn{2}{c}{6.61}\\
$\Delta_{R=Gray}$ & & &  \multicolumn{2}{c|}{\textbf{3.83}} & & & \multicolumn{2}{c|}{\textbf{4.08}} & & & \multicolumn{2}{c|}{\textbf{4.59}} & & & \multicolumn{2}{c|}{\textbf{3.88}} & & & \multicolumn{2}{c}{\textbf{4.71}}\\
\bottomrule
\end{tabular}
\vspace{0.2cm}
\caption{Average PSNR of the investigated SR models evaluated on Set5. R=RGB and R=Gray indicate the channels of the representative image for INT8 PTQ. $\Delta$ is the PSNR difference between clip and no-clip models. Note that we construct 800 different RD's with a single image for each network, then average the resulting INT8 networks' PSNR values.}
\label{tab:PSNR_representative_G_RGB}
\end{table}

\textbf{\ul{Do clipped activations really affect PTQ?}} Table~\ref{tab:PSNR_representative_G_RGB} clearly shows that removing the clip without any additional care for INT8 PTQ hinders the PSNR drastically ($\Delta$ values), but increases the latency significantly as shown in Table~\ref{tab:speedy}. Therefore, it is worthwhile to eliminate the clipped activation without sacrificing the visual quality since PTQ inherently applies clipping to the network by using the data statistics in RD anyway. 

\textit{Hence, now the issue boils down to how to characterize RD such that it can replace the clipped activation layer. }

\begin{table}[ht!]
\setlength{\tabcolsep}{1.5pt}
\centering
\fontsize{5pt}{5pt}\selectfont
\begin{tabular}{@{}rrcccccccccccccccccccc@{}}
\toprule
\multicolumn{1}{c}{\multirow{2}{*}{RD Size}} & \multicolumn{1}{c}{\multirow{2}{*}{RD}}                                                                                              & \multicolumn{4}{c}{\textbf{XCAT}}                                 & \multicolumn{4}{c}{\textbf{ABPN}}                                 & \multicolumn{4}{c}{\textbf{ESPCN}}                                & \multicolumn{4}{c}{\textbf{FSRCNN}}                               & \multicolumn{4}{c}{\textbf{RFDNNet}}            \\ \cmidrule(l){3-22} 
\multicolumn{1}{c}{}                         & \multicolumn{1}{c}{}                                                                                                                 &all RD&min&max&avg                   &all RD&min&max&avg                  &all RD&min&max&avg                   &all RD&min&max&avg                  &all RD&min&max&avg \\ \cmidrule(r){1-22}
\multicolumn{1}{r|}{5} & \multicolumn{1}{r|}{\textit{Good} images} &
\textbf{32.84} & 32.57 & 32.85 & \multicolumn{1}{l|}{32.72} &
\textbf{33.11} & 32.86 & 33.11 & \multicolumn{1}{l|}{33.00} &
\textbf{31.44} & 30.86 & 31.82 & \multicolumn{1}{l|}{31.53} &
\textbf{32.03} & 30.41 & 32.01 & \multicolumn{1}{l|}{31.40} &
\textbf{32.28} & 29.79 & 32.16 & 31.54 
\\ \midrule

\multicolumn{1}{r|}{5} & \multicolumn{1}{r|}{\textit{Bad} images} &
\textbf{22.95} & 22.95 & 25.70 & \multicolumn{1}{l|}{24.05} &
\textbf{20.93} & 20.93 & 24.84 & \multicolumn{1}{l|}{23.41} &
\textbf{20.20} & 20.20 & 23.43 & \multicolumn{1}{l|}{21.94} &
\textbf{21.48} & 21.47 & 24.85 & \multicolumn{1}{l|}{23.49} &
\textbf{21.60} & 21.72 & 24.09 & 23.16 
\\ \midrule

\multicolumn{1}{r|}{\multirow{3}{*}{5}}     & \multicolumn{1}{r|}{\multirow{3}{*}{\begin{tabular}[c]{@{}r@{}}Random selections\\ from\\ \textit{good} and \textit{bad}\end{tabular}}} &
\textbf{23.69} & 23.47 & 32.85 & \multicolumn{1}{l|}{29.50} &
\textbf{24.22} & 24.39 & 33.11 & \multicolumn{1}{l|}{29.67} &
\textbf{21.74} & 21.83 & 31.79 & \multicolumn{1}{l|}{28.04} &
\textbf{22.40} & 21.83 & 31.79 & \multicolumn{1}{l|}{28.04} &
\textbf{23.08} & 21.83 & 31.79 & 28.04 
\\
\multicolumn{1}{r|}{} & \multicolumn{1}{r|}{}&                       \textbf{22.95} & 22.95 & 31.71 & \multicolumn{1}{l|}{27.22} &
\textbf{20.93} & 20.93 & 33.04 & \multicolumn{1}{l|}{26.87} &
\textbf{20.20} & 20.20 & 31.82 & \multicolumn{1}{l|}{25.46} &
\textbf{21.47} & 21.47 & 31.80 & \multicolumn{1}{l|}{26.60} &
\textbf{21.71} & 21.72 & 31.89 & 26.11 
\\
\multicolumn{1}{r|}{} & \multicolumn{1}{r|}{}&                       \textbf{22.95} & 22.95 & 32.86 & \multicolumn{1}{l|}{28.95} &
\textbf{20.93} & 20.93 & 33.04 & \multicolumn{1}{l|}{28.83} &
\textbf{20.20} & 20.20 & 31.79 & \multicolumn{1}{l|}{27.21} &
\textbf{21.47} & 21.47 & 31.49 & \multicolumn{1}{l|}{27.77} &
\textbf{21.72} & 21.72 & 32.04 & 27.89 
\\
\midrule
\multicolumn{1}{r|}{10} & \multicolumn{1}{r|}{\textit{Good} + \textit{bad}} &
\textbf{22.95} & 22.95 & 32.85 & \multicolumn{1}{l|}{28.38} &
\textbf{20.93} & 20.93 & 33.11 & \multicolumn{1}{l|}{28.20} &
\textbf{20.20} & 20.20 & 31.82 & \multicolumn{1}{l|}{26.73} &
\textbf{21.48} & 21.47 & 32.02 & \multicolumn{1}{l|}{27.44} &
\textbf{21.60} & 21.72 & 32.16 & 27.35 
\\ \midrule
\multicolumn{1}{r|}{100} & \multicolumn{1}{r|}{DIV2K training images} &                                       
\textbf{22.08} & 22.08 & 32.85 & \multicolumn{1}{l|}{28.19} &
\textbf{21.61} & 21.61 & 33.06 & \multicolumn{1}{l|}{28.13} &
\textbf{19.87} & 19.87 & 31.79 & \multicolumn{1}{l|}{26.04} &
\textbf{19.40} & 31.96 & 19.39 & \multicolumn{1}{l|}{26.89} &
\textbf{20.75} & 20.72 & 32.04 & 26.83 
\\ \bottomrule
\end{tabular}
\vspace{0.2cm}
\caption{The effect of the size of RD on Set5 PSNR for no-clip INT8 models. \textit{all RD} denotes the PSNR where the model is quantized with all images in RD. We also pick each image from the RD's stated with their sizes, create new RD's with size one, perform PTQ, evaluate the INT8 models, and state the resulting PSNR's \textit{min}, \textit{max}, and \textit{avg} values.  Notice that the existence of \textit{bad} images inside RD considerably decreases the overall performance.}
\label{tab:num_of_imgs_in_RD}
\end{table}

\textbf{\ul{Does the number of images in RD affect the PTQ quality?}} In our study, \textit{good} RD images are the ones that have the smallest PSNR difference between their clipped INT8 and no-clip INT8 models, and vice versa for the \textit{bad} ones. We experimented with different-sized RD's and concluded that including only a single image in the RD during PTQ is the most straightforward and tractable solution. This is because if there are multiple images in the RD, \textit{bad} representative images are observed to overpower the others and lead to a \textit{worse} quantized model in the end, too. This can be observed from the results in Table~\ref{tab:num_of_imgs_in_RD}.

\textit{From now on, we continue with RD's only consisting of a single image. RD with size one and representative image (RI) denote the same concept unless specified otherwise.}

\textbf{\ul{Is PTQ dependent on the number of channels of RI?}} Table~\ref{tab:PSNR_representative_G_RGB} reveals that the PSNR differences between INT8 clipped and no-clip models ($\Delta$) are minimized when the RI is chosen as grayscale instead of RGB. Hence, it would be more logical to work on grayscale representative images and try to make $\Delta$ as small as possible by altering them. Note that this is surprisingly valid when the input images to the network are RGB, but the RI is grayscale.

\textit{From now on, we continue with RIs only consisting of a grayscale image.}

\begin{figure}[ht!]
\centering
\includegraphics[scale=0.2, trim={0 0 0 2cm}, clip]{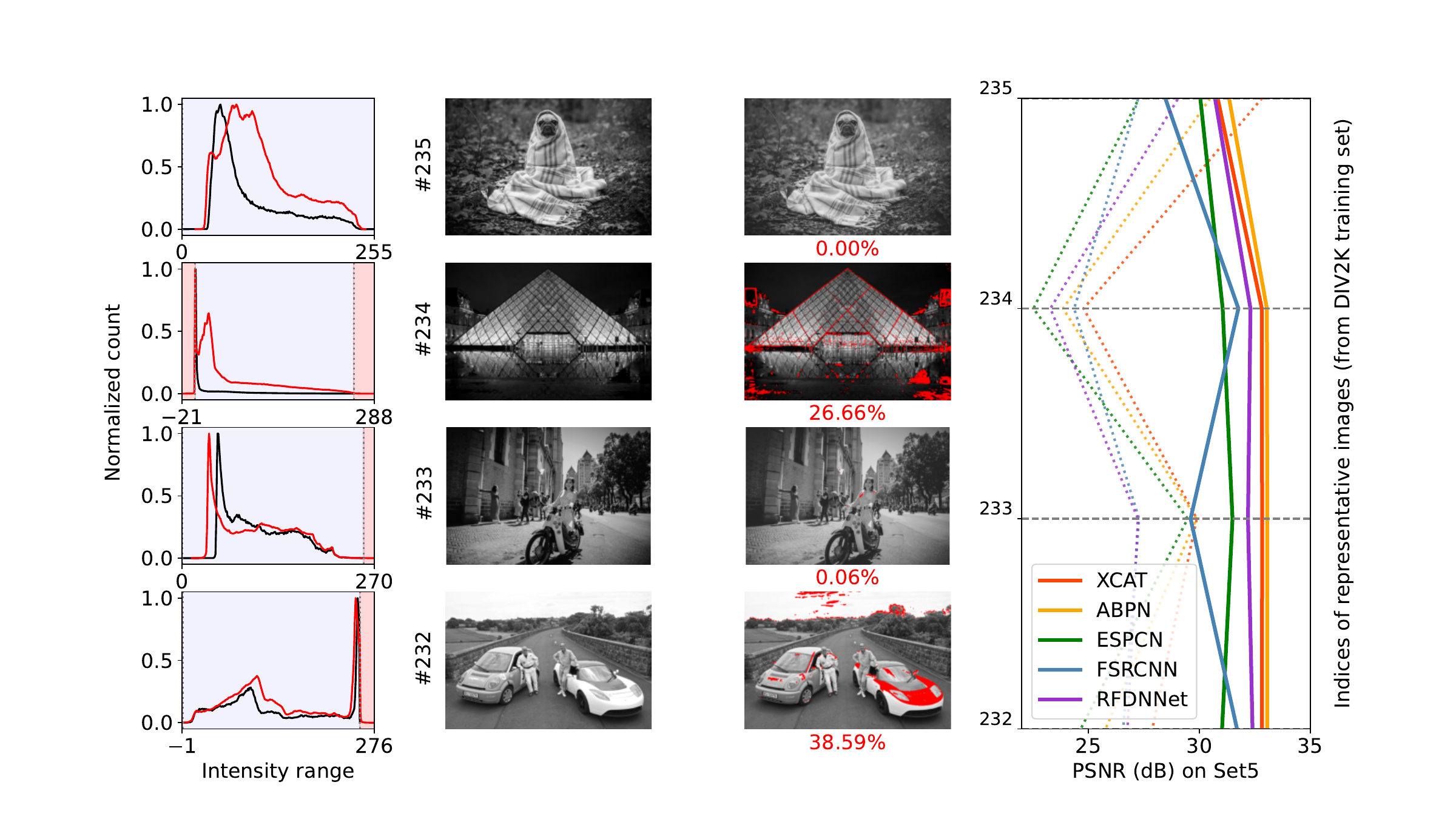}
\caption{PSNR scores of some of the quantized SR models with different representative images (right). Dashed lines are no-clip INT8 models, and solid lines are clipped INT8. The response of corresponding FP32 models when the selected representative images are given as inputs (middle). Outlier pixels are marked with red, stated with their percentages. The histograms of the input (black) and output data (red) of the FP32 model (left), where the red areas in the histograms match with the distribution of outlier pixels.}
\label{fig:good_RD_preliminary}
\end{figure}

\textbf{\ul{What can indicate \textit{good} representative images?}} To find a common point in those \textit{good} images, we construct 800 grayscale RI's, perform PTQ on five different no-clip \& clipped FP32 SR models, and evaluate the resulting INT8 models on Set5. The findings are given in Figure~\ref{fig:good_RD_preliminary}.

One interesting finding is that the response of the FP32 network to the corresponding representative image yields a correlation with $\Delta_{R=Gray}$ in Table~\ref{tab:PSNR_representative_G_RGB}. To see it in action, consider image \#234 in DIV2K training set in Figure~\ref{fig:good_RD_preliminary}. The corresponding models when RI is set as \#234 have significant PSNR gaps between their clipped INT8 and the no-clip INT8 models (hence \#234 is a \textit{bad} image). Then, when \#234 is given to no-clip FP32 models (XCAT is considered for demonstration without loss of generality), 26.6\% of the output pixels are observed to be out-of-range (>255, <0). Note that the more out-of-range the FP32 outputs (red areas in Figure~\ref{fig:good_RD_preliminary}), the more it is inclined towards yielding a lower PSNR. The opposite happens for \#235, where even some no-clip INT8 models perform better than their clipped counterparts (negative $\Delta_{R=Gray}$, a very \textit{good} image for RI). 

\textit{Hence, we conclude that the number of outliers in the FP32 model's response to the corresponding RI gives us a measure of whether the image is good or bad, correlated with $\Delta_{R=Gray}$. Therefore, if we reduce the outliers, we may also decrease $\Delta_{R=Gray}$ and increase the PTQ stability. \textbf{This also means that we can understand whether a representative image would be considered as \textit{good} or \textit{bad} without performing PTQ. Then, if we correctly identify and use good representative images, we can omit the clipped activation layer without any loss in the INT8 model.}}

Even though Figure~\ref{fig:good_RD_preliminary} only shows a very small subset of possible RIs, we show our proposal's positive effect and generalizability in Figure~\ref{fig:circle_plots} later on. It is also worth noting that we also investigated other image characteristics (like their histograms) before passing it through the no-clip FP32 network but did not capture relevant correlations with the PTQ quality (Table~\ref{tab:histogram_matching}, \ref{tab:histogram_rotate}). We also experimented with masking out the outlier sections with other healthy regions from the same image but did not achieve satisfactory results.

\begin{table}[ht!]
\setlength{\tabcolsep}{2pt}
\centering
\fontsize{7pt}{7pt}\selectfont
\begin{tabular}{@{}c|ccc|ccc|ccc@{}}
\toprule 
{\textbf{Models}} & \multicolumn{1}{c|}{\textbf{\#105$\rightarrow$\#508}} & \multicolumn{1}{c|}{\textbf{\#508}} & \textbf{\#105} & \multicolumn{1}{c|}{\textbf{\#001$\rightarrow$\#216}} & \multicolumn{1}{c|}{\textbf{\#216}} & \textbf{\#001} & \multicolumn{1}{c|}{\textbf{\#416$\rightarrow$\#176}} & \multicolumn{1}{c|}{\textbf{\#176}} & \textbf{\#416} \\ \midrule
XCAT                             & \multicolumn{1}{c|}{22.472}           & \multicolumn{1}{c|}{22.956}         & 32.860         & \multicolumn{1}{c|}{22.982}           & \multicolumn{1}{c|}{23.471}         & 32.850         & \multicolumn{1}{c|}{29.403}           & \multicolumn{1}{c|}{23.764}         & 32.716         \\ 
ABPN                             & \multicolumn{1}{c|}{21.980}           & \multicolumn{1}{c|}{20.933}         & 33.026         & \multicolumn{1}{c|}{22.445}           & \multicolumn{1}{c|}{24.849}         & 33.115         & \multicolumn{1}{c|}{29.418}           & \multicolumn{1}{c|}{24.195}         & 32.864         \\ 
ESPCN                            & \multicolumn{1}{c|}{23.720}           & \multicolumn{1}{c|}{20.208}         & 31.544         & \multicolumn{1}{c|}{24.382}           & \multicolumn{1}{c|}{21.838}         & 31.628         & \multicolumn{1}{c|}{28.856}           & \multicolumn{1}{c|}{21.654}         & 31.820         \\
FSRCNN                           & \multicolumn{1}{c|}{21.564}           & \multicolumn{1}{c|}{21.475}         & 31.492         & \multicolumn{1}{c|}{21.879}           & \multicolumn{1}{c|}{24.497}         & 32.019         & \multicolumn{1}{c|}{19.394}           & \multicolumn{1}{c|}{24.233}         & 31.801         \\ 
RFDNNet                          & \multicolumn{1}{c|}{23.955}           & \multicolumn{1}{c|}{21.723}         & 32.044         & \multicolumn{1}{c|}{20.463}           & \multicolumn{1}{c|}{23.454}         & 32.165         & \multicolumn{1}{c|}{26.044}           & \multicolumn{1}{c|}{24.098}         & 31.893         \\ \bottomrule
\end{tabular}
\vspace{0.2cm}
\caption{PSNR on Set5 for no-clip INT8 models, where histogram matching is applied to RI images. \# denote the image index in DIV2K training set. When \textit{good} image histograms are mapped onto the \textit{bad} ones, an improvement is not observed, showing that the histogram is not a correlation factor.}
\label{tab:histogram_matching}
\end{table}

\begin{table}[ht!]
\vspace{-0.2cm}
\setlength{\tabcolsep}{2pt}
\centering
\fontsize{6pt}{6pt}\selectfont
\begin{tabular}{c|cccc|cccc|cccc|cccc|cccc}
\toprule
& \multicolumn{4}{c|}{XCAT}                                                                    & \multicolumn{4}{c|}{ABPN}                                                                    & \multicolumn{4}{c|}{ESPCN}                                                                   & \multicolumn{4}{c|}{FSRCNN}                                                                  & \multicolumn{4}{c}{RFDNNet}                                                                 \\ \cmidrule{2-21} 
& {0°} & {90°} & {180°} & {270°} &  
{0°} & {90°} & {180°} & {270°} &  
{0°} & {90°} & {180°} & {270°} &  
{0°} & {90°} & {180°} & {270°} &  
{0°} & {90°} & {180°} & {270°} 
\\ \cmidrule{2-21} 
\multicolumn{1}{c}{\#105} & \multicolumn{1}{|c}{32.86} & \multicolumn{1}{c}{32.83} & \multicolumn{1}{c}{32.87} & 32.86 & \multicolumn{1}{c}{33.03} & \multicolumn{1}{c}{33.05} & \multicolumn{1}{c}{32.78} & 32.99 & \multicolumn{1}{c}{31.54} & \multicolumn{1}{c}{31.53} & \multicolumn{1}{c}{31.36} & 31.64 & \multicolumn{1}{c}{31.49} & \multicolumn{1}{c}{31.43} & \multicolumn{1}{c}{31.38} & 31.51 & \multicolumn{1}{c}{32.04} & \multicolumn{1}{c}{31.59} & \multicolumn{1}{c}{32.20} & 32.14 \\ 
\multicolumn{1}{c|}{\#508} & \multicolumn{1}{c}{22.96} & \multicolumn{1}{c}{23.25} & \multicolumn{1}{c}{23.21} & 22.64 & \multicolumn{1}{c}{20.93} & \multicolumn{1}{c}{20.61} & \multicolumn{1}{c}{21.24} & 18.79 & \multicolumn{1}{c}{20.21} & \multicolumn{1}{c}{20.07} & \multicolumn{1}{c}{19.29} & 20.02 & \multicolumn{1}{c}{21.48} & \multicolumn{1}{c}{21.52} & \multicolumn{1}{c}{21.62} & 20.89 & \multicolumn{1}{c}{21.72} & \multicolumn{1}{c}{21.63} & \multicolumn{1}{c}{22.07} & 21.90 \\ \bottomrule
\end{tabular}
\vspace{0.2cm}
\caption{PSNR on Set5 for no-clip INT8 models, where given RI's are rotated. Despite the rotation not changing the histogram, the output PNSR's are different.}
\label{tab:histogram_rotate}
\end{table}

\textbf{\ul{How can be systematically find \textit{good} representative images and improve the \textit{bad} ones?}} The procedure of finding \textit{good} \& \textit{bad} images based on the previous observations is easy: if the image yields \textit{low} the number of outlier pixels after the no-clip FP32 model, then quantizing that no-clip FP32 model with that image set as the representative will yield a well-performing no-clip INT8 model, and vice versa for the \textit{bad} one. What about improving the \textit{bad} RIs? At this stage, we propose our RI augmentation pipeline (CFQP) in Section~\ref{sec:methodology}.  

\section{Methods and Pipeline}
\label{sec:methodology}
This section presents our two methods for choosing \textit{good} RI's and improving the \textit{bad} ones. Ultimately, we combine those methods to create CFQP.

\begin{figure}[h!]
	\centering
	\includegraphics[scale=0.48]{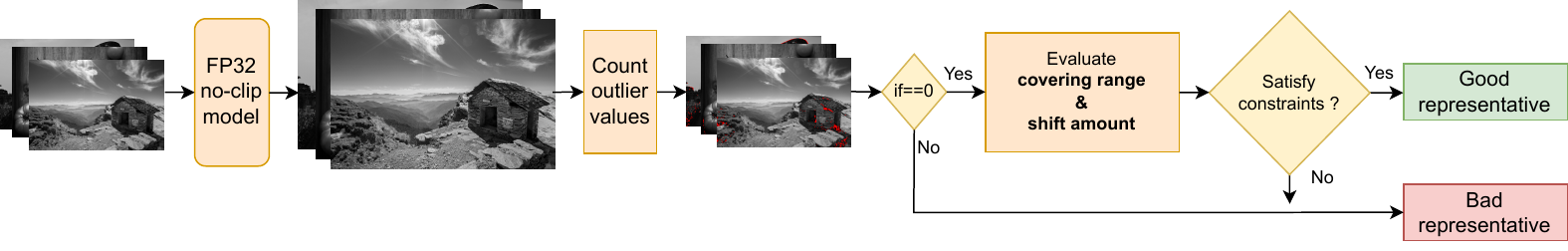}
	\caption{The procedure of selecting \textit{good} representative images.}
    \label{fig:method1}
\end{figure}

\textbf{\ul{Method \#1: Good Representative Selection:}}
Method \#1 given in Figure~\ref{fig:method1} aims to separate \textit{good} RI from \textit{bad}. Based on the previous observations, we put three constraints that a \textit{good} RI needs to satisfy after passing through the FP32 network: it has zero outliers, has to have \textit{min-max} values \textit{close to} 0-255, and its \textit{min-max} values need not deviate \textit{much} from the \textit{min-max} of the original RI. All algorithmic design parameters are discussed in Section~\ref{sec:preliminary_observations} and observed to be valid for most of the considered data. The first and second constraints stem from avoiding dividing the pixel values into 255 bins during PTQ, therefore eliminating decreasing the quantization quality due to having wider bins than required. The third constraint ensures that the FP32 model does not change or shift the pixel distribution of RI, as this will also cause the inferred images from the INT8 model to have color shifts, making the network unpractical. We empirically tuned the maximum \textit{min-max} pixel difference to 5, deviation pixel difference to 5, maximum out-of-range pixels threshold as 25, and maximum shift thresholds as {10}, respectively, by using 10 images from the DIV2K train set. These hyperparameters are easy to adjust, and our choices are observed to work on nearly all DIV2K train set (800 images). 

Table~\ref{tab:final_improvement} shows the PSNR values of these selected images for different models. As seen, the images selected by Method \#1 are confirmed to be the \textit{good} ones. No-clip with good RI mostly performs better than the clipped cases. The next step is to augment the other images such that they also perform \textit{good}.

\textbf{\ul{Method \#2: Enhancement of Bad Representatives:}} The goal of this method is to bring the pixel of FP32 outputs of the \textit{bad} RIs into the 0-255 range (i.e., decrease the count of outlier values) without distorting the textural information significantly. As stated in Section~\ref{sec:preliminary_observations}, modifying the input RI histograms does not yield satisfactory results. Hence, we have investigated several other image processing techniques and discovered that blurring the image achieves our goal for this method. This can be also verified from Figure~\ref{fig:good_RD_preliminary} as well. The models usually overshoot the high-frequency content of the image (such as edges, corners, etc.), and low-pass filtering them keeps the overall structure of the original image and eliminates the high-frequency content, reducing the generation of outliers. Therefore, we used several blurring techniques in CFQP as well (Algorithm~\ref{alg:alg1}, \ref{alg:alg2}, and \ref{alg:alg3}).

{
\begin{framed}
\captionsetup{hypcap=false}
\noindent\begin{minipage}{\textwidth}
  \begin{minipage}{.25\textwidth}
  \vspace{-8.6mm}
  \footnotesize{
    \captionof{algorithm}{GB}
    \vspace{2mm}
    \label{alg:alg1}
    \begin{algorithmic}
    \IF{$ON(Image)$$\neq 0$}
    \STATE $Image \Leftarrow GB(Image)$
    \ENDIF
    \end{algorithmic}}
  \end{minipage}
  \begin{minipage}{.365\textwidth}
  \vspace{-2mm}
  \footnotesize{
    \captionof{algorithm}{LIBB}\vspace{2mm}
    \label{alg:alg2}
    \begin{algorithmic}
    \WHILE{$ON(Image) \neq 0$ \OR $IN \neq Thr$ }
    \STATE $Image \Leftarrow LIBB(Image)$
    \STATE $IN \Leftarrow IN+1 $
    \ENDWHILE
    \end{algorithmic}}
  \end{minipage}
  \begin{minipage}{.365\textwidth}
  \footnotesize{
    \vspace{-2mm}
    \captionof{algorithm}{LIPB}\vspace{2mm}
    \label{alg:alg3}
    \begin{algorithmic}
    \WHILE{$ON(Image) \neq 0$ \OR $IN \neq Thr$ }
    \STATE $Image \Leftarrow LIPB(Image)$
    \STATE $IN \Leftarrow IN+1 $
    \ENDWHILE
    \end{algorithmic}}
  \end{minipage}
  \vspace{-1mm}
  \label{alg:algorithms}
\end{minipage}
\end{framed}
}
\textit{ON} is outlier number after the image is inferred in the no-clip FP32 model, \textit{IN} is iteration number, \textit{Thr} is threshold for iteration number, \textit{GB} is Global Blur, \textit{LIBB} is Local Iterative Box Blur, and \textit{LIPB} is Local Iterative Point Blur. For all, we experimented with different blur parameters  \textit{(3x3, 5x5, 7x7 kernels)} and observed that 3x3 works the best.

\textbf{Global Blur (GB, Algorithm~\ref{alg:alg1}):} A \textit{3x3} homogeneous blur filter is applied on the whole image. We observed a significant drop in outlier numbers for most bad representatives. After we used that blurred image as RI, we observed a dramatic increase in the PSNR values of INT8 models, given in Section~\ref{sec:exp_results}. Extra blurring also causes a significant drop in the INT8 visual quality output, causing the textural information to disappear. Therefore, GB is not applied iteratively.

\textbf{Local Iterative Box Blur (LIBB, Algorithm~\ref{alg:alg2}):} A \textit{3x3} homogeneous filter is applied to a rectangular region in the image. This box is defined by four lines, including outlier pixels with the highest height, lowest height, highest width, and lowest width. This blur drops outlier numbers in the box region without touching the unproblematic pixels. However, it may create unwanted new outlier pixels at the box's border. To address this issue, we apply the filter iteratively to reduce outlier numbers further. We limit the number of iterations using a threshold to avoid excessive blurring (hence decreasing the PSNR score).

\textbf{Local Iterative Point Blur (LIPB, Algorithm~\ref{alg:alg3}):} LIPB approach is similar to the LIBB. We observed that large box areas in LIBB may not always be effective in improving the PSNR score. Therefore, we used another mechanism to reduce the blurred regions in an extreme case. We masked outlier pixels on the FP32 model's output image (HR) and blurred on corresponding points in the input (LR) image so we would not touch other unproblematic regions. It can be considered a trade-off; averaging is done among very few pixels where most are outliers, but the blurring of unproblematic regions is minimized.

\subsection{Proposed Pipeline Structure}
We combine Method 1 and 2 to construct our CFQP. Method 1 decides whether the image is \textit{good} enough for RD. If not, the image undergoes the augmentation procedure. Method 2 applies three distinct blur operations to the image. The image quality as an RI is measured by checking the outlier numbers, determined by the response of the FP32 model. It is an interactive structure such that Method 2 benefits from the FP32 model output to measure outlier numbers, which will determine the RI to quantize the same FP32 model. 
After the blurs, Method 2 selects the best image based on the minimum outlier number. This selected image becomes the new RI for quantization. Consequently, models have as many RIs as the training dataset, and each one can be employed to quantize the model. To pick among them, one may use methods like RANSAC among the possible selections. The pipeline is presented in Section~\ref{sec:exp_results}, and the effect of the pipeline can be examined visually by comparing Figure~\ref{fig:good_RD_preliminary} with Figure~\ref{fig:R1_selections_final}.

\begin{figure}[ht!]
    \centering
    \includegraphics[scale=0.135]{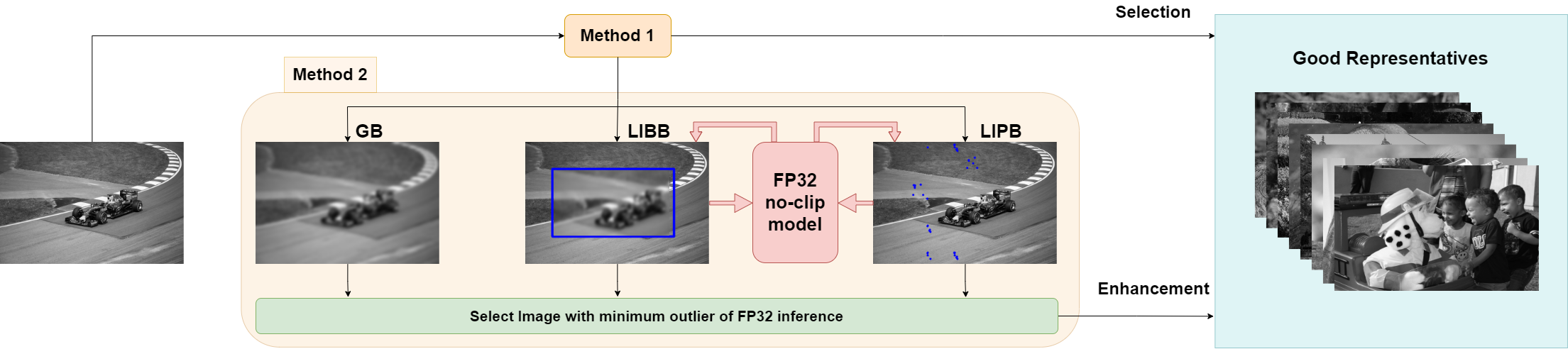}
    \vspace{3mm}
    \caption{Workflow of CFQP.}
    \label{fig:pipeline}
\end{figure}

\section{Results}
\label{sec:exp_results}
\begin{figure}[ht!]
    \centering
    \includegraphics[scale=0.25, trim={2cm 1cm 3cm 3cm}, clip]{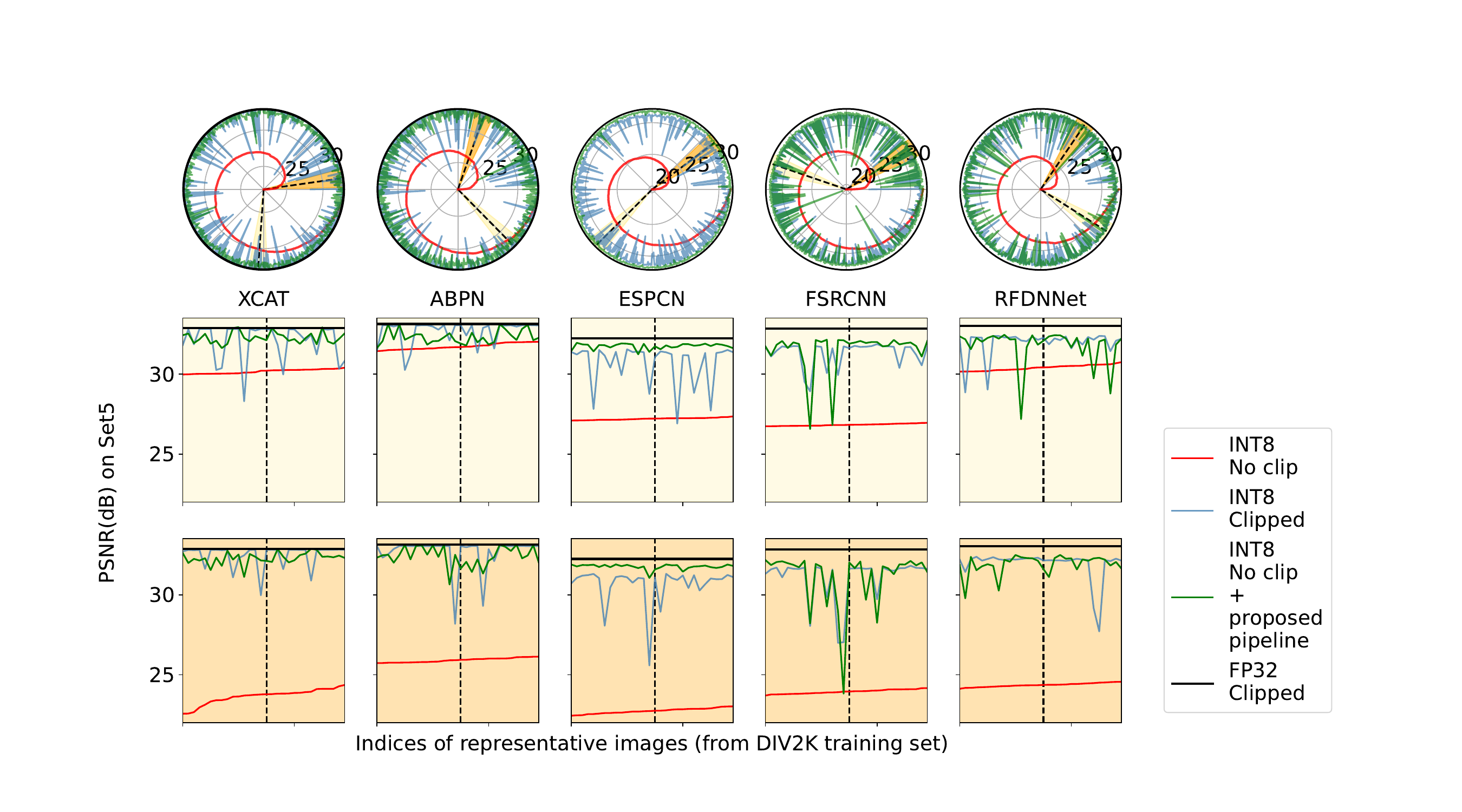}
    \caption{PSNR on Set5 (radial \& y-axis) for different models, where each model is quantized with different RI selections (angular \& x-axis). Our proposed pipeline applied on the no-clip INT8 (green) outperforms the no-clip INT8 by directly using RI images (red). For clarity, we sort INT8 no-clip PSNR values (red) and use the same sorted indices for the rest of the data for consistency.}
    \label{fig:circle_plots}
\end{figure}

\begin{figure}[ht!]
    \centering
    \includegraphics[scale=0.5]{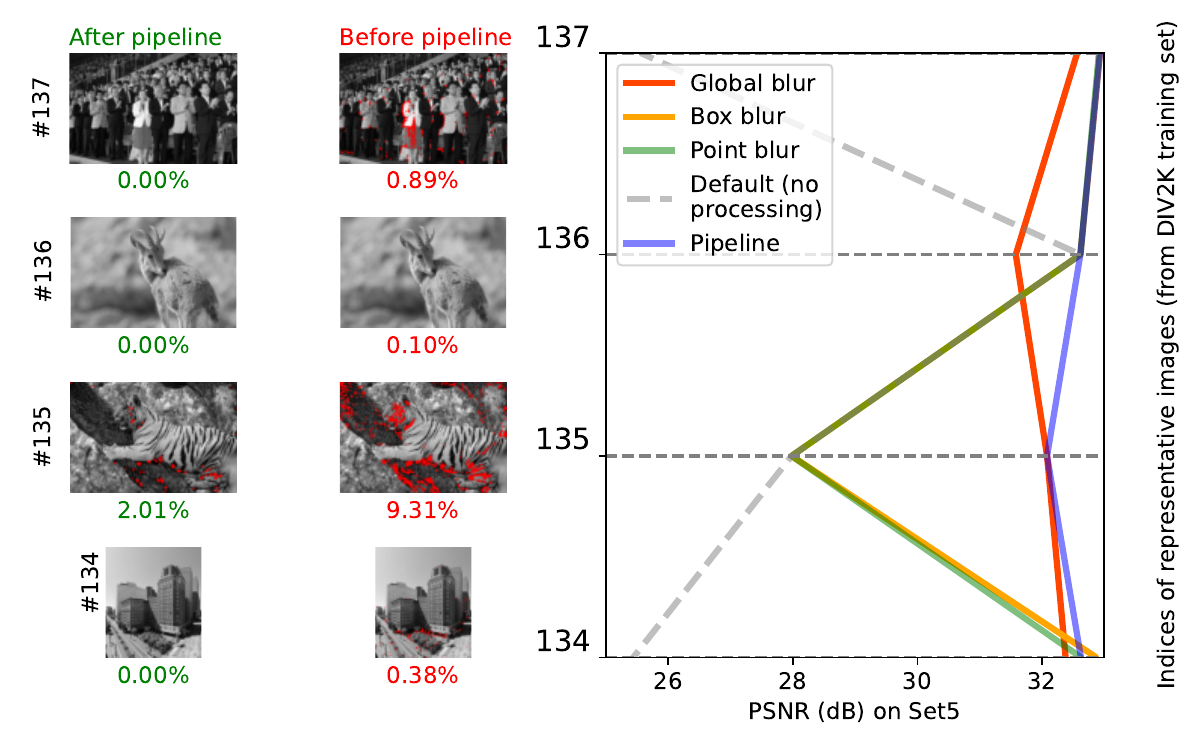}
    \caption{PSNR on Set5 for no-clip XCAT for the possible RI selections. The best augmentation methods for the corresponding RI and the FP32 output responses are also provided. Our proposed pipeline gives us the best augmentation selections for RI on average.}
    \label{fig:R1_selections_final}
    \vspace{-0.5cm}
\end{figure}

We provide the results of our proposed work in Table~\ref{tab:speedy}, \ref{tab:final_improvement}, and Figure~\ref{fig:visuals},~\ref{fig:circle_plots}, \ref{fig:R1_selections_final}. Our extensive experiments reveal that CFQP performs as intended and as discussed in previous sections.
\begin{table}[ht!]
\setlength{\tabcolsep}{1.5pt}
\centering
\fontsize{7pt}{7pt}\selectfont
\begin{tabular}{@{}r|ccc|ccc|ccc|ccc|ccc@{}}
\toprule
\multicolumn{1}{l}{}                 & \multicolumn{3}{c}{XCAT}                                                               & \multicolumn{3}{c}{ABPN}                                                               & \multicolumn{3}{c}{ESPCN}                                                              & \multicolumn{3}{c}{FSRCNN}                                                             & \multicolumn{3}{c}{RFDNNet}                                                            \\ \cmidrule(lr){2-4} \cmidrule(lr){5-7} \cmidrule(lr){8-10} \cmidrule(lr){11-13} \cmidrule(lr){14-16}
& \cmark & \xmark & $\uparrow$ & \cmark & \xmark & $\uparrow$ & \cmark & \xmark & $\uparrow$ & \cmark & \xmark & $\uparrow$ & \cmark & \xmark & $\uparrow$ \\ \midrule
CPU                                  & 458                        & 379                        & 20.8\%                        & 779                        & 715                        & 9.0\%                         & 328                        & 213                        & \textbf{54.0\%}                        & 1610                       & 1564                       & 2.9\%                         & 904                        & 810                        & 11.6\%                        \\
TFLite GPU                           & 460                        & 445                        & 3.4\%                         & 790                        & 781                        & 1.2\%                         & 342                        & 339                        & 0.9\%                         & 667                        & 554                        & 20.4\%                        & 484                        & 465                        & 4.1\%                         \\
Android NNAPI                        & 451                        & 416                        & 8.4\%                         & 254                        & 197                        & 28.9\%                        & 187                        & 155                        & 20.6\%                        & 309                        & 275                        & 12.4\%                        & 600                        & 545                        & 10.1\%                        \\
Qualcomm Hexagon                     & 377                        & 338                        & 11.5\%                        & 178                        & 150                        & 18.7\%                        & 113                        & 91                         & 24.2\%                        & 1207                       & 1206                       & 0\%                           & 519                        & 498                        & 4.2\%                         \\
Qualcomm QNN GPU & 502                        & 375                        & 33.9\%                        & 772                        & 728                        & 6.0\%                         & 322                        & 213                        & 51.2\%               & 1605                       & 1552                       & 3.4\%                         & 917                        & 836                        & 9.7\%                         \\ \bottomrule
\end{tabular}
\vspace{0.2cm}
\caption{Average runtime (ms) on different backends for clipped (\cmark) INT8 models, no-clip INT8 (\xmark) models, and the improvement over no-clip INT8 models ($\uparrow$). We achieve 14.86\% $\uparrow$ on average for all models. The input tensor dimension is 640x360, and the inferences are repeated 20 times.~\cite{benchmark} is used for inference on Xiaomi RedMi Note 8 mobile phone.}
\label{tab:speedy}
\vspace{-0.5cm}
\end{table}

\begin{table}[ht!]
\setlength{\tabcolsep}{1.5pt}
\centering
\fontsize{7pt}{7pt}\selectfont
\begin{tabular}{@{}cccccc@{}}
\toprule
  \multirow{2}{*}{Models} & \multicolumn{4}{c}{Test sets}                              \\ \cmidrule{2-5} 
& \multicolumn{1}{c|}{Set5} & \multicolumn{1}{c|}{Set14} & \multicolumn{1}{c|}{BSD100} & \multicolumn{1}{c}{Urban100}  \\ \midrule
XCAT & 32.98/32.86 & 29.41/29.36 & 28.38/28.34 & 26.30/26.20 \\
 ABPN & 33.18/33.00 & 29.51/29.52 & 28.51/28.43 & 26.56/26.42 \\
 ESPCN & 32.23/32.72 & 28.89/28.85 & 28.04/28.43 & 25.56/25.86  \\
 FSRCNN & 32.83/32.72 & 29.25/29.31 & 28.27/28.24 & 26.01/25.96 \\
 RFDNNet & 32.97/32.27 & 29.33/29.35 & 28.39/28.67 & 26.24/26.04 \\
 \bottomrule
\end{tabular}
\vspace{0.2cm}
\caption{PSNR of FP32 models on different train sets (DIV2K/Flickr).}
\label{tab:tab_flickr}
\vspace{-0.5cm}
\end{table}

\begin{figure}[ht!]
    \centering
    \addtolength{\tabcolsep}{-5pt}
    \fontsize{7pt}{7pt}\selectfont

    \begin{tabular}{cccccc}
    $\Delta$ & INT8 & \textbf{INT8*} & $\Delta$ & INT8 & \textbf{INT8*} \\
    \includegraphics[width=0.1\linewidth]{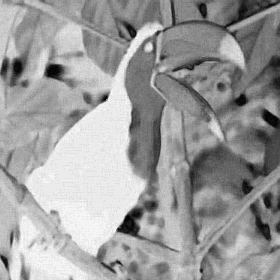}&
    \includegraphics[width=0.1\linewidth]{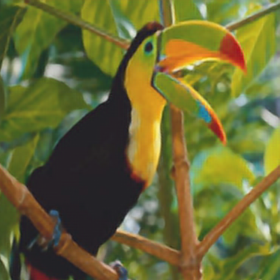}&
    \includegraphics[width=0.1\linewidth]{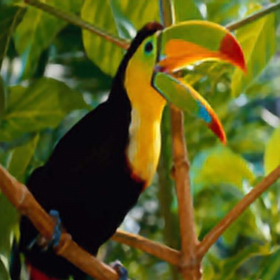}&
    \includegraphics[width=0.1\linewidth]{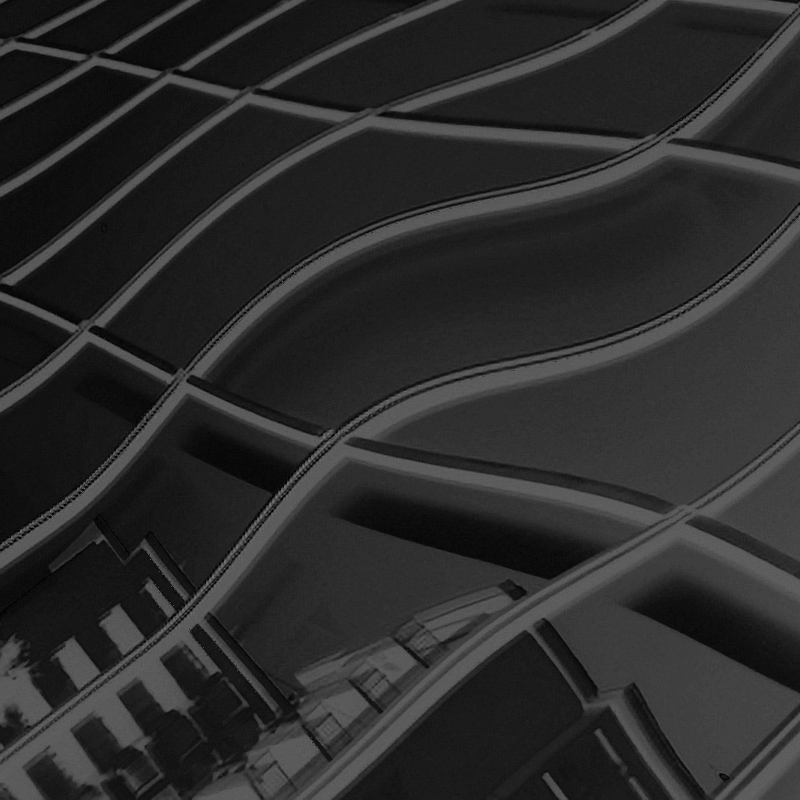}&
    \includegraphics[width=0.1\linewidth]{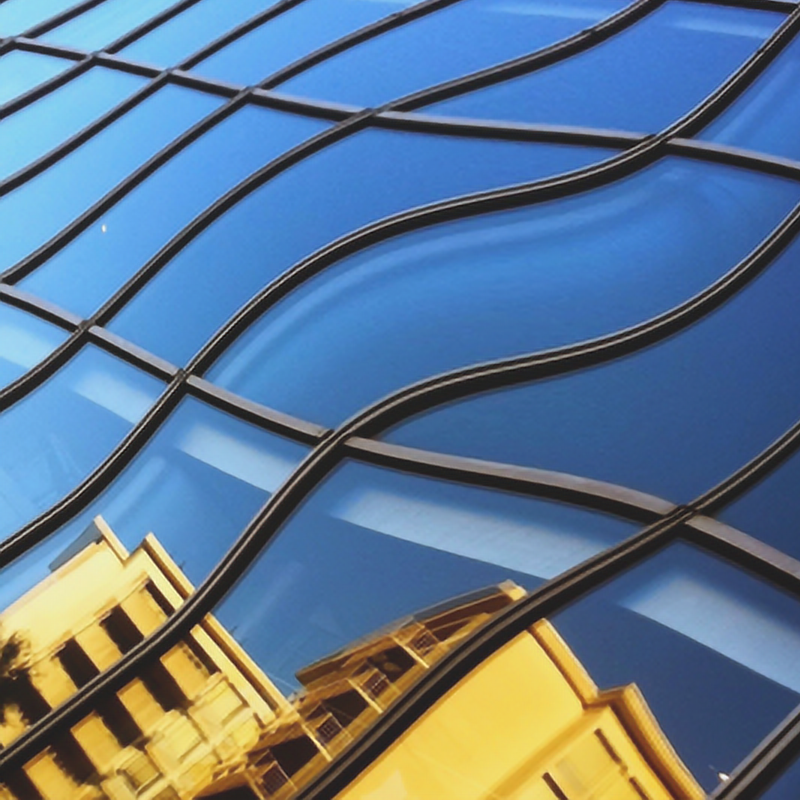}&
    \includegraphics[width=0.1\linewidth]{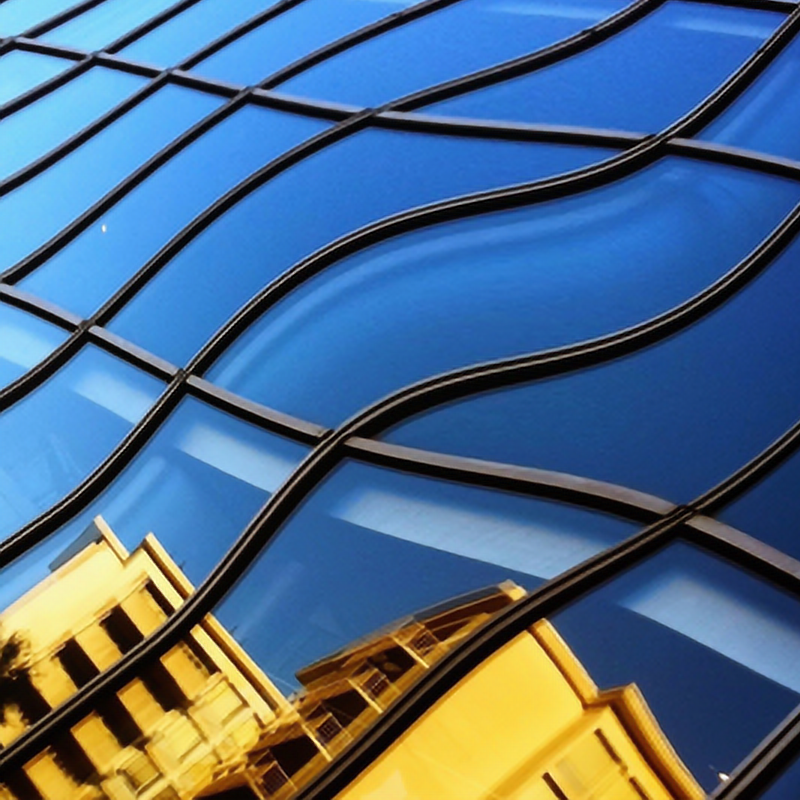}\\
      & 21.21/0.896 & 33.40/0.958 &   & 24.64/0.914 & 33.00/0.948 \\
    \end{tabular}
    \vspace{0.2cm}
    \caption{No-clip XCAT PSNR/SSIM results. * whose RI is undergone our pipeline yield better results. Notice the color inconsistencies due to unaltered \textit{bad} RI (DIV2K train \#736).}
    \label{fig:visuals}
    \vspace{-0.5cm}
\end{figure}

\begin{table}[ht!]
\setlength{\tabcolsep}{1.5pt}
\centering
\fontsize{6pt}{6pt}\selectfont
\begin{tabular}{@{}lc|cc|cc|c@{}}
\toprule
\multirow{2}{*}{Test} & \multirow{2}{*}{Models} & \multicolumn{2}{c|}{Clipped}                  & \multicolumn{3}{c}{No-Clip}              \\ \cmidrule{3-7}& & \multicolumn{1}{c|}{M1} & \cancel{M1} & \multicolumn{1}{c|}{M1} & \cancel{M1} & $\text{\cancel{M1}}\rightarrow\text{M2}$ \\ \midrule
\multirow{5}{*}{{\rotatebox[origin=c]{90}{Set5}}}&\multicolumn{1}{|c|}{XCAT}                                                                                                  & \multicolumn{1}{c|}{\color{Green}{32.329}  \textit{(15)}}          & \color{Red}32.310  \textit{(785)} & \multicolumn{1}{c|}{\color{Green}32.640\textit{ \textit{(15)}}}        & \color{Red}{28.409}  \textit{(785)} & \textbf{\color{RoyalBlue}32.346  \textit{(785)}} \\ 
&\multicolumn{1}{|c|}{ABPN}                                                                                                & \multicolumn{1}{c|}{\color{Green}32.492  \textit{(15)}}          & \color{Red}32.526  \textit{(785)} & \multicolumn{1}{c|}{\color{Green}32.975  \textit{(15)}}        & \color{Red}{28.360}  \textit{(785)}  & \textbf{\color{RoyalBlue}32.276  \textit{(785)}} \\ 
&\multicolumn{1}{|c|}{ESPCN}                                                                                                 & \multicolumn{1}{c|}{\color{Green}31.139  \textit{(8)}}          & \color{Red}30.855  \textit{(792)} & \multicolumn{1}{c|}{\color{Green}31.501  \textit{(8)}}        &  \color{Red}{26.213}  \textit{(792)} & \textbf{\color{RoyalBlue}31.758  \textit{(792)}} \\ 
&\multicolumn{1}{|c|}{FSRCNN}                                                                                             & \multicolumn{1}{c|}{\color{Green}31.232  \textit{(13)}}          & \color{Red}31.124  \textit{(787)} & \multicolumn{1}{c|}{\color{Green}31.529  \textit{(13)}}         &  \color{Red}{27.178}  \textit{(787)} & \textbf{\color{RoyalBlue}31.439  \textit{(787)}} \\ 
&\multicolumn{1}{|c|}{RFDNNet}                                                                                            & \multicolumn{1}{c|}{\color{Green}32.167 \textit{(7)}}          & \color{Red}31.766  \textit{(793)} & \multicolumn{1}{c|}{\color{Green}32.079 \textit{(7)}}        &  \color{Red}{27.012}  \textit{(793)} & \textbf{\color{RoyalBlue}31.798  \textit{(793)}} \\  \midrule

\multirow{5}{*}{{\rotatebox[origin=c]{90}{Set14}}}&\multicolumn{1}{|c|}{XCAT}                                                                                                  & \multicolumn{1}{c|}{\color{Green}{28.956}  \textit{(15)}}          & \color{Red}28.890  \textit{(785)} & \multicolumn{1}{c|}{\color{Green}29.131\textit{ \textit{(15)}}}        & \color{Red}{26.516}  \textit{(785)} & \textbf{\color{RoyalBlue}28.910  \textit{(785)}} \\ 
&\multicolumn{1}{|c|}{ABPN}                                                                                                & \multicolumn{1}{c|}{\color{Green}29.042  \textit{(15)}}          & \color{Red}28.956  \textit{(785)} & \multicolumn{1}{c|}{\color{Green}29.360  \textit{(15)}}        & \color{Red}{26.380}  \textit{(785)}  & \textbf{\color{RoyalBlue}28.906  \textit{(785)}} \\ 
&\multicolumn{1}{|c|}{ESPCN}                                                                                                 & \multicolumn{1}{c|}{\color{Green}28.239  \textit{(8)}}          & \color{Red}28.616  \textit{(792)} & \multicolumn{1}{c|}{\color{Green}28.478  \textit{(8)}}        &  \color{Red}{25.089}  \textit{(792)} & \textbf{\color{RoyalBlue}28.649  \textit{(792)}} \\ 
&\multicolumn{1}{|c|}{FSRCNN}                                                                                             & \multicolumn{1}{c|}{\color{Green}28.096  \textit{(13)}}          & \color{Red}28.149  \textit{(787)} & \multicolumn{1}{c|}{\color{Green}28.241  \textit{(13)}}         &  \color{Red}{25.539}  \textit{(787)} & \textbf{\color{RoyalBlue}28.271  \textit{(787)}} \\ 
&\multicolumn{1}{|c|}{RFDNNet}                                                                                            & \multicolumn{1}{c|}{\color{Green}29.058 \textit{(7)}}          & \color{Red}28.906  \textit{(793)} & \multicolumn{1}{c|}{\color{Green}28.877 \textit{(7)}}        &  \color{Red}{25.605}  \textit{(793)} & \textbf{\color{RoyalBlue}28.748  \textit{(793)}} \\ \midrule

\multirow{5}{*}{{\rotatebox[origin=c]{90}{BSD100}}}&\multicolumn{1}{|c|}{XCAT}                                                                                                  & \multicolumn{1}{c|}{\color{Green}{28.094}  \textit{(15)}}          & \color{Red}28.126  \textit{(785)} & \multicolumn{1}{c|}{\color{Green}28.216\textit{ \textit{(15)}}}        & \color{Red}{25.907}  \textit{(785)} & \textbf{\color{RoyalBlue}28.126  \textit{(785)}} \\ 
&\multicolumn{1}{|c|}{ABPN}                                                                                                & \multicolumn{1}{c|}{\color{Green}28.165  \textit{(15)}}          & \color{Red}28.196  \textit{(785)} & \multicolumn{1}{c|}{\color{Green}28.408  \textit{(15)}}        & \color{Red}{26.069}  \textit{(785)}  & \textbf{\color{RoyalBlue}28.150  \textit{(775)}} \\ 
&\multicolumn{1}{|c|}{ESPCN}                                                                                                 & \multicolumn{1}{c|}{\color{Green}27.494  \textit{(8)}}          & \color{Red}27.821  \textit{(792)} & \multicolumn{1}{c|}{\color{Green}27.684  \textit{(8)}}        &  \color{Red}{24.960}  \textit{(792)} & \textbf{\color{RoyalBlue}27.843  \textit{(792)}} \\ 
&\multicolumn{1}{|c|}{FSRCNN}                                                                                             & \multicolumn{1}{c|}{\color{Green}27.611  \textit{(13)}}          & \color{Red}27.736  \textit{(787)} & \multicolumn{1}{c|}{\color{Green}27.723  \textit{(13)}}         &  \color{Red}{25.452}  \textit{(787)} & \textbf{\color{RoyalBlue}27.785  \textit{(787)}} \\ 
&\multicolumn{1}{|c|}{RFDNNet}                                                                                            & \multicolumn{1}{c|}{\color{Green}28.231 \textit{(7)}}          & \color{Red}28.149  \textit{(793)} & \multicolumn{1}{c|}{\color{Green}28.165 \textit{(7)}}        &  \color{Red}{25.459}  \textit{(793)} & \textbf{\color{RoyalBlue}28.073  \textit{(793)}} \\  \midrule

\multirow{5}{*}{{\rotatebox[origin=c]{90}{Urban100}}}&\multicolumn{1}{|c|}{XCAT}                                                                                                  & \multicolumn{1}{c|}{\color{Green}{25.964}  \textit{(15)}}          & \color{Red}25.967  \textit{(785)} & \multicolumn{1}{c|}{\color{Green}26.137\textit{ \textit{(15)}}}        & \color{Red}{24.328}  \textit{(785)} & \textbf{\color{RoyalBlue}25.845  \textit{(775)}} \\ 
&\multicolumn{1}{|c|}{ABPN}                                                                                                & \multicolumn{1}{c|}{\color{Green}26.197  \textit{(15)}}          & \color{Red}26.261  \textit{(785)} & \multicolumn{1}{c|}{\color{Green}26.440  \textit{(15)}}        & \color{Red}{24.318}  \textit{(785)}  & \textbf{\color{RoyalBlue}25.943  \textit{(775)}} \\ 
&\multicolumn{1}{|c|}{ESPCN}                                                                                                 & \multicolumn{1}{c|}{\color{Green}25.211  \textit{(8)}}          & \color{Red}25.104  \textit{(792)} & \multicolumn{1}{c|}{\color{Green}25.349  \textit{(8)}}        &  \color{Red}{23.056}  \textit{(792)} & \textbf{\color{RoyalBlue}25.448  \textit{(792)}} \\ 
&\multicolumn{1}{|c|}{FSRCNN}                                                                                             & \multicolumn{1}{c|}{\color{Green}25.410  \textit{(13)}}          & \color{Red}25.444  \textit{(787)} & \multicolumn{1}{c|}{\color{Green}25.420  \textit{(13)}}         &  \color{Red}{23.537}  \textit{(787)} & \textbf{\color{RoyalBlue}25.397  \textit{(787)}} \\ 
&\multicolumn{1}{|c|}{RFDNNet}                                                                                            & \multicolumn{1}{c|}{\color{Green}26.046 \textit{(7)}}          & \color{Red}25.922  \textit{(793)} & \multicolumn{1}{c|}{\color{Green}25.971 \textit{(7)}}        &  \color{Red}{23.619}  \textit{(793)} & \textbf{\color{RoyalBlue}25.815  \textit{(793)}} \\ \bottomrule
\end{tabular}
\vspace{0.2cm}
\vspace{0.2cm}
\caption{Average of differently quantized models {\color{Green}with the images (good) in Method 1 (M1)}, {\color{Red}images which are rejected (bad) by Method 1 ($\text{\cancel{M1}}$) }, and {\color{NavyBlue}the bad images improved by our Method 2 ($\text{\cancel{M1}}\rightarrow\text{M2}$)} on 4 different validation sets. Each PNSR entry is the average of \textit{(n)} differently quantized models, where their representative images are chosen from DIV2K training set (800).}
\label{tab:final_improvement}
\vspace{-0.5cm}
\end{table}

\section{Conclusion}
In this work, we showed the importance of representative image selection for PTQ. Specifically, we have presented a novel RI selection and enhancement pipeline (CFQP) for PTQ of no-clip SR networks. It is based on reducing outlier values in images with the selected image augmentation methods, making every image in a training dataset a candidate for RI. Results reveal that our proposed method on no-clip INT8 models outperforms clipped INT8 model counterparts regarding visual quality and provides a faster inference due to clipping removal. For future work, we aim to work on creating good representative images using generative networks.

\label{sec:conc}


\begin{thebibliography}{20}
\providecommand{\natexlab}[1]{#1}
\providecommand{\url}[1]{\texttt{#1}}
\expandafter\ifx\csname urlstyle\endcsname\relax
  \providecommand{\doi}[1]{doi: #1}\else
  \providecommand{\doi}{doi: \begingroup \urlstyle{rm}\Url}\fi

\bibitem[Agustsson and Timofte(2017)]{div2k}
Eirikur Agustsson and Radu Timofte.
\newblock Ntire 2017 challenge on single image super-resolution: Dataset and
  study.
\newblock In \emph{The IEEE Conference on Computer Vision and Pattern
  Recognition (CVPR) Workshops}, July 2017.

\bibitem[Ayazoglu(2021)]{xlsr}
Mustafa Ayazoglu.
\newblock Extremely lightweight quantization robust real-time single-image
  super resolution for mobile devices.
\newblock In \emph{Proceedings of the IEEE/CVF Conference on Computer Vision
  and Pattern Recognition (CVPR) Workshops}, pages 2472--2479, June 2021.

\bibitem[Ayazoglu and Bilecen(2023)]{xcat}
Mustafa Ayazoglu and Bahri~Batuhan Bilecen.
\newblock Xcat - lightweight quantized single image super-resolution using
  heterogeneous group convolutions and cross concatenation.
\newblock In \emph{Computer Vision -- ECCV 2022 Workshops}, pages 475--488,
  Cham, 2023. Springer Nature Switzerland.

\bibitem[Bevilacqua et~al.(2012)Bevilacqua, Roumy, Guillemot, and line
  Alberi~Morel]{set5}
Marco Bevilacqua, Aline Roumy, Christine Guillemot, and Marie line
  Alberi~Morel.
\newblock Low-complexity single-image super-resolution based on nonnegative
  neighbor embedding.
\newblock In \emph{Proceedings of the British Machine Vision Conference}, pages
  135.1--135.10. BMVA Press, 2012.

\bibitem[Choi et~al.(2018)Choi, Wang, Venkataramani, Chuang, Srinivasan, and
  Gopalakrishnan]{PACT}
Jungwook Choi, Zhuo Wang, Swagath Venkataramani, Pierce~I{-}Jen Chuang,
  Vijayalakshmi Srinivasan, and Kailash Gopalakrishnan.
\newblock {PACT:} parameterized clipping activation for quantized neural
  networks.
\newblock \emph{CoRR}, 2018.

\bibitem[Conde et~al.(2023)Conde, Zamfir, Timofte, et~al.]{conde2023ntire_rtsr}
Marcos~V Conde, Eduard Zamfir, Radu Timofte, et~al.
\newblock Efficient deep models for real-time 4k image super-resolution. ntire
  2023 benchmark and report.
\newblock In \emph{Proceedings of the IEEE/CVF Conference on Computer Vision
  and Pattern Recognition}, 2023.

\bibitem[Dong et~al.(2016)Dong, Loy, and Tang]{fsrcnn}
Chao Dong, Chen~Change Loy, and Xiaoou Tang.
\newblock Accelerating the super-resolution convolutional neural network.
\newblock In Bastian Leibe, Jiri Matas, Nicu Sebe, and Max Welling, editors,
  \emph{Computer Vision -- ECCV 2016}, pages 391--407, Cham, 2016. Springer
  International Publishing.

\bibitem[Du et~al.(2021)Du, Liu, Tang, and Wu]{abpn}
Zongcai Du, Jie Liu, Jie Tang, and Gangshan Wu.
\newblock Anchor-based plain net for mobile image super-resolution.
\newblock In \emph{Proceedings of the IEEE/CVF Conference on Computer Vision
  and Pattern Recognition (CVPR) Workshops}, pages 2494--2502, June 2021.

\bibitem[Haichao~Yu(2021)]{cross-domain}
Humphrey~Shi Haichao~Yu, Linjie~Yang.
\newblock Is in-domain data really needed? a pilot study on cross-domain
  calibration for network quantization.
\newblock In \emph{Proceedings of the IEEE/CVF Conference on Computer Vision
  and Pattern Recognition}, 2021.

\bibitem[Ignatov and Timofte(2023)]{aim2022}
Andrey Ignatov and Radu Timofte.
\newblock Efficient and accurate quantized image super-resolution on mobile
  npus, mobile ai {\&} aim 2022 challenge: Report.
\newblock In \emph{Computer Vision -- ECCV 2022 Workshops}, pages 92--129,
  Cham, 2023. Springer Nature Switzerland.

\bibitem[Ignatov et~al.(2019)Ignatov, Timofte, Kulik, Yang, Wang, Baum, Wu, Xu,
  and Van~Gool]{benchmark}
Andrey Ignatov, Radu Timofte, Andrei Kulik, Seungsoo Yang, Ke~Wang, Felix Baum,
  Max Wu, Lirong Xu, and Luc Van~Gool.
\newblock Ai benchmark: All about deep learning on smartphones in 2019, 10
  2019.

\bibitem[Ignatov et~al.(2021)Ignatov, Timofte, Denna, and Younes]{mai2021}
Andrey Ignatov, Radu Timofte, Maurizio Denna, and Abdel Younes.
\newblock Real-time quantized image super-resolution on mobile npus, mobile ai
  2021 challenge: Report.
\newblock In \emph{Proceedings of the IEEE/CVF Conference on Computer Vision
  and Pattern Recognition (CVPR) Workshops}, pages 2525--2534, June 2021.

\bibitem[Li et~al.(2022)Li, Zhang, Timofte, and Gool]{esr_ntire_2022}
Yawei Li, Kai Zhang, Radu Timofte, and Van Gool.
\newblock Ntire 2022 challenge on efficient super-resolution: Methods and
  results.
\newblock In \emph{Proceedings of the IEEE/CVF Conference on Computer Vision
  and Pattern Recognition (CVPR) Workshops}, pages 1062--1102, June 2022.

\bibitem[Liu et~al.(2020)Liu, Tang, and Wu]{rfdnnet}
Jie Liu, Jie Tang, and Gangshan Wu.
\newblock Residual feature distillation network for lightweight image
  super-resolution, 2020.

\bibitem[Ruonan~Yu(2023)]{dataset-distillation}
Xinchao~Wang Ruonan~Yu, Songhua~Liu.
\newblock Dataset distillation: A comprehensive review.
\newblock In \emph{IEEE Transactions on Pattern Analysis and Machine
  Intelligence}, 2023.

\bibitem[Shi et~al.(2016)Shi, Caballero, Huszar, Totz, Aitken, Bishop,
  Rueckert, and Wang]{espcn}
Wenzhe Shi, Jose Caballero, Ferenc Huszar, Johannes Totz, Andrew~P. Aitken, Rob
  Bishop, Daniel Rueckert, and Zehan Wang.
\newblock Real-time single image and video super-resolution using an efficient
  sub-pixel convolutional neural network.
\newblock In \emph{Proceedings of the IEEE Conference on Computer Vision and
  Pattern Recognition (CVPR)}, June 2016.

\bibitem[Yuan et~al.(2023)Yuan, Liu, Wu, Yang, Wu, Sun, Liu, Wang, and
  Wu]{benchmark_PTQ}
Zhihang Yuan, Jiawei Liu, Jiaxiang Wu, Dawei Yang, Qiang Wu, Guangyu Sun, Wenyu
  Liu, Xinggang Wang, and Bingzhe Wu.
\newblock Benchmarking the reliability of post-training quantization: a
  particular focus on worst-case performance, 2023.

\bibitem[Zamfir et~al.(2023)Zamfir, Conde, and Timofte]{zamfir2023rtsr}
Eduard Zamfir, Marcos~V Conde, and Radu Timofte.
\newblock Towards real-time 4k image super-resolution.
\newblock In \emph{Proceedings of the IEEE/CVF Conference on Computer Vision
  and Pattern Recognition}, 2023.

\bibitem[Zhang et~al.(2022)Zhang, Gao, Fan, Zhao, Yang, and Yan]{selectQ}
Zhao Zhang, Yangcheng Gao, Jicong Fan, Zhongqiu Zhao, Yi~Yang, and Shuicheng
  Yan.
\newblock {SelectQ: Calibration Data Selection for Post-Training Quantization},
  11 2022.

\bibitem[Zhuo et~al.(2022)Zhuo, Chen, Ramakrishnan, Chen, Feng, Lin, Zhang, and
  Shen]{LPQ_tinyML}
Shaojie Zhuo, Hongyu Chen, Ramchalam~Kinattinkara Ramakrishnan, Tommy Chen,
  Chen Feng, Yicheng Lin, Parker Zhang, and Liang Shen.
\newblock An empirical study of low precision quantization for tinyml, 2022.

\end{thebibliography}
\end{document}